\documentclass[5p]{elsarticle}
\usepackage{amsmath,amsfonts,amssymb}
\usepackage{array}
\usepackage[caption=false,font=normalsize,labelfont=sf,textfont=sf]{subfig}
\usepackage{textcomp}
\usepackage{stfloats}
\usepackage{url}
\usepackage{verbatim}
\usepackage{graphicx}
\usepackage{setspace}
\usepackage{cite}
\usepackage{hyperref}
\usepackage[utf8]{inputenc}
\usepackage{times}  
\usepackage{helvet}  
\usepackage{courier}  
\usepackage[ruled,lined,linesnumbered]{algorithm2e}
\DontPrintSemicolon
\SetKwComment{tcc}{$\triangleright$~}{}
\SetCommentSty{normalfont}
\SetKwInput{KwRequire}{Require}
\SetNlSty{}{}{:}
\usepackage{paralist}
\usepackage[most]{tcolorbox}
\usepackage{todonotes}
\usepackage[capitalise]{cleveref}
\usepackage[legacycolonsymbols]{mathtools}
\usepackage{booktabs}
\usepackage{wrapfig}
\usepackage{flushend}
\graphicspath{ {./fig/} }

\begin{document}

\title{Reinforcement Learning with Elastic Time Steps\tnoteref{t1}}
\tnotetext[t1]{This work was partly presented at the Finding the Frame Workshop in August 2024.}

\author[1]{Dong Wang}
\ead{dong-1.wang@polymtl.ca}
\author[2]{Giovanni Beltrame\corref{cor1}}
\ead{giovanni.beltrame@polymtl.ca}
\cortext[cor1]{Corresponding author}

\affiliation[1,2]{organization={Polytechnique Montr\'eal},
	addressline={2500, chemin de Polytechnique},
	postcode={H3T 1J4},
	city={Montr\'eal (Qu\'ebec)},
	country={Canada}}

\begin{abstract}
	Traditional Reinforcement Learning (RL) policies are typically implemented
	with fixed control rates, often disregarding the impact of control rate
	selection. This can lead to inefficiencies as the optimal control rate varies
	with task requirements. We present the Multi-Objective Soft Elastic
	Actor-Critic (MOSEAC), an off-policy actor-critic algorithm that uses elastic
	time steps to dynamically adjust the control frequency. This technique
	minimizes computational resources by selecting the lowest viable frequency. We
	demonstrate that MOSEAC converges and produces stable policies at the theoretical
	level, and validate our findings in a real-time 3D racing game. MOSEAC
	significantly outperformed other variable time step approaches in terms of
	energy efficiency and task effectiveness. Additionally, MOSEAC shown
	faster and more stable training, showcasing its potential for real-world RL
	applications in robotics.
\end{abstract}



\maketitle

\section{Introduction}
Model-free deep Reinforcement Learning (DRL) has
shown notable value in diverse domains, from video games
\citep{silver2016mastering, tmrl} to robotic control \citep{haarnoja2018soft2,
	wurman2022outracing}. For example, Sony's autonomous racing cars achieved
remarkable results with fixed training frequencies of 10 Hz and 60 Hz
\citep{wurman2022outracing}.

However, suboptimal fixed control rates present limitations, leading to 
excessive caution, wasted computational resources, risky behavior, and 
compromised control.

Variable Time Step Reinforcement Learning (VTS-RL) has been introduced to 
address these issues in traditional RL by utilizing reactive programming 
principles \citep{majumdar2021paracosm, bregu2016reactive}. VTS-RL performs 
control actions only when necessary, reducing computational load and enabling 
variable action durations. For example, in robotic manipulation, VTS-RL allows 
a robot arm to dynamically adjust its control frequency, utilizing lower frequencies 
for simple tasks and higher frequencies for complex maneuvers or delicate 
handling \citep{karimi2023dynamic}.

Two notable VTS-RL algorithms are Soft Elastic Actor-Critic 
(SEAC) \citep{wang2024deployable} and Continuous-Time Continuous-Options (CTCO) 
\citep{karimi2023dynamic}. CTCO supports continuous-time decision-making with flexible 
option durations, enhancing exploration and robustness. However, it requires 
tuning multiple hyperparameters, such as radial basis functions (RBFs) and 
time-related parameters $\tau$ for its adaptive discount factor $\gamma$, making 
tuning complex in certain environments.

SEAC incorporates reward components related to task energy (number of actions) 
and task time, making it effective in time-constrained environments. Despite its 
advantages, SEAC requires careful tuning of hyperparameters to balance task, 
energy, and time costs to ensure optimal performance. The sensitivity of both 
SEAC and CTCO to hyperparameter settings presents a challenge for users aiming 
to fully exploit their capabilities.

We recently proposed Multi-Objective Soft Elastic Actor-Critic
(MOSEAC~\citep{wang2024moseac}), which reduced the dimension of hyperparameters
and the algorithm’s dependence on hyperparameters by dynamically adjusting the
hyperparameters corresponding to the reward structure.

We identify shortcomings in our previous work, where we proposed adapting
$\alpha_m$ by monitoring reward trends. However, in some tasks, reward trends
are not always stable, posing a risk of reward explosion. To address this, we
introduce an upper limit $\alpha_{max}$ for $\alpha_m$. We provid pseudocode
for this improvement and established a corresponding Lyapunov stability function
\citep{lyapunov1992general} to demonstrate the stability (convergence) of the new
algorithm.

We evaluate MOSEAC in a racing game (Ubisoft TrackMania \citep{trackmania}),
comparing it against CTCO~\citep{karimi2023dynamic}, SEAC~\citep{wang2024deployable}
and Soft Actor-Critic (SAC~\citep{haarnoja2018soft1}). Our results demonstrate
MOSEAC’s improved training speed, stability, and efficiency.
Our key contributions are:
\begin{compactenum}
	\item \textbf{Algorithm Enhancement and Convergence Proof:} We introduce an
	upper limit $\alpha_{\text{max}}$ on the parameter $\alpha_m$ to prevent
	reward explosion, ensuring the stability and convergence of the MOSEAC
	algorithm. This enhancement is validated through a Lyapunov model, providing a
	proof of convergence and demonstrating the efficacy of the new adjustment
	mechanism.
	
	
	\item \textbf{Enhanced Training Efficiency:} We demonstrate that \\
	MOSEAC achieves faster
	and more stable training than CTCO, highlighting its practical benefits and
	applicability in real-world scenarios.
\end{compactenum}

The paper is organized as follows. Section~\ref{sec:related} describes the 
current research status of variable time step RL. Section~\ref{sec:overview} 
introduces MOSEAC with its pseudocode and Lyapunov model. Section~\ref{sec:env} 
describes the test environments. Section~\ref{sec:results} presents the 
simulation parameters and results. Finally, Section~\ref{sec:conclusions} 
concludes the paper.

\section{Related Work}
\label{sec:related}

Fixed control rates in RL often lead to inefficiencies. Research by
\citep{wang2024deployable} shows that suboptimal fixed rates can cause excessive
caution or risky behavior, wasting resources and compromising control. Control
rates significantly impact continuous control systems beyond computational
demands. Some studies \citep{amin2020locally,park2021time} indicate that
high control rates can degrade RL performance, while low rates hinder
adaptability in complex scenarios.

\citet{sharma2017learning} proposed learning to repeat actions to
mimic dynamic control rates, but this technique does not change the control
frequency or reduce computational demands. Few studies have explored repetitive
behaviors in real-world scenarios, such as those by \citet{metelli2020control} 
and \citet{lee2020reinforcement}. \citet{chen2021varlenmarl} introduced variable 
``control rates'' utilizing actions like ``sleep,'' but still involved 
fixed-frequency checks.

\citet{cui2022reinforcement} applied the Lyapunov model to verify the
stability of RL algorithms and addressed handling the dynamics of power systems
over time, although it still uses a fixed frequency to scan system states.

Introducing time steps of variable duration allows a robotic system to better
adapt to its task and environment by adjusting control actions based on the
system's current conditions, addressing the nonlinearity
and time-variant dynamics typical in robotics \citep{shin1985minimum}.
This adaptability ensures optimal performance through efficient resource
utilization and effective response to varying conditions
\citep{bainomugisha2013survey, bregu2016reactive}.


Lyapunov models provide robust stability guarantees by employing a scalar
function that decreases over time, ensuring system convergence to a desired
equilibrium. By incorporating Lyapunov stability into our RL
system, we ensure that the learning process remains stable, thereby
preventing erratic behavior and potential system
failure~\citep{cui2022reinforcement, chow2018lyapunov}.

\section{Multi-Objective Soft Elastic Actor and Critic}
\label{sec:overview}

Our algorithm builds upon previous work \citep{wang2024moseac}, combining SEAC's
hyperparameters for balancing task, energy, and time rewards through a simple
multiplication technique, and applying adaptive adjustments to the remaining
hyperparameters. A key improvement is the introduction of an upper limit for the
hyperparameter $\alpha_m$. Below is an overview of the MOSEAC algorithm. This
overview emphasizes the critical aspects of MOSEAC without delving into the
detailed definitions. The reader can refer to~\citep{wang2024moseac} for details.


The reward in MOSEAC is:
\begin{equation}\label{eq:reward_policy}
R = \alpha_{m} R_t R_{\tau} - \alpha_{\varepsilon}
\end{equation}
where \(R_t\) is the task reward, $R_{\tau}$ is a time-dependent term,
$\alpha_{m}$ is a weighting factor to modulate reward magnitude, and
$\alpha_{\varepsilon}=0.2 \cdot \left(1 - \frac{1}{1 + e^{-\alpha_m +
		1}}\right)$ is a penalty parameter applied at each time step to reduce
unnecessary actions.

To automatically set $\alpha_{m}$ to an optimal value, we dynamically adjust it
during training. This adjustment mitigates convergence issues, specifically the
problem of sparse rewards caused by suboptimal settings of $\alpha_{m}$. MOSEAC
increases $\alpha_{m}$ (and decrease $\alpha_{\varepsilon}$) if the average reward
is declining.

We introduce a hyperparameter $\psi$ to dynamically adjust $\alpha_m$ based 
on observed trends in task rewards:
\begin{equation}
\alpha_{m} = 
\begin{cases} 
\alpha_{m} + \psi & \text{if } \alpha_{m} < \alpha_{max} \\
\alpha_{m} = \alpha_{max} & \text{otherwise} 
\end{cases}
\end{equation}

In this work, we add $\alpha_{max}$ to ensure convergence and prevent reward
explosion. 


Algorithm \ref{algo:SEAC} presents the pseudocode for MOSEAC. MOSEAC extends the 
SAC algorithm by incorporating action duration $D$ into the action policy set, 
enabling it to predict both the action and its duration simultaneously. The 
reward is computed utilizing \autoref{eq:reward_policy} and is continuously monitored. 
If the reward trend declines, $\alpha_m$ increases linearly at a rate of $\psi$, 
but does not exceed $\alpha_{max}$. The action and critic networks are periodically 
updated, similar to the SAC algorithm, based on these preprocessed rewards.

\begin{algorithm}[ht]\label{algo:SEAC}
	\SetAlgoLined
	\KwRequire{a policy $\pi$ with a set of parameters $\theta$, $\theta^{'}$, 
		critic parameters $\phi$, $\phi^{'}$, variable time step environment 
		model $\Omega$, learning-rate $\lambda_p$, $\lambda_q$, reward buffer 
		$\beta_r$, replay buffer $\beta$.}
	
	Initialization $i = 0$, $t_i = 0$, $\beta_r = 0$, observe $S_0$\\
	\While{$t_i \leq t_{max}$}{
		\For{$i \leq k_{length} \vee Not \, Done$}{
			$A_i, D_i = \pi_{\theta}(S_{i})$ \\
			$S_{i+1}, R_i = \Omega(A_i, D_i)$ \\ 
			
			$i \leftarrow i+1$
		}
		$\beta_{r} \leftarrow 1/i \times \sum_{0}^{i} R_i$ \\ 
		$\beta \leftarrow S_{0 \sim i}, \, A_{0 \sim i}, \, D_{0 \sim i}, \, 
		R_{0 \sim i}, \, S_{1 \sim i+1}$ \\
		$i = 0$\\
		$t_{i} \leftarrow t_{i} + 1$ \\ 
		\If{$t_{i} \geq k_{init} \quad \& \quad t_{i} \mid k_{update}$}{
			$Sample \, S, \, A, \, D, \, R, \, S^{'} from (\beta)$ \\
			$\phi \leftarrow \phi - \lambda_q\nabla_{\delta}\mathcal{L}_{Q}
			(\phi, \, S, \, A, \, D, \, R, \, S^{'})$ \\ \hfill {$\rightarrow$ 
				critic update}\\
			$\theta \leftarrow \theta - \lambda_p\nabla_{\theta}\mathcal{L}_{
				\pi}(\theta, \, S, \, A, \, D, \, \phi)$ \\ \hfill 
			{$\rightarrow$ actor update}\\
			\If{$k_R(\beta_r)$}{
				
				$\alpha_{m} = \alpha_{m} + \psi$  \hfill $\text{if} \alpha_{m} 
				< \alpha_{max}$ \\
				$\text{Or}$ $\alpha_{m} = \alpha_{max}$  \hfill 
				$\text{otherwise}$ \\
				
				$\alpha_{\varepsilon} \leftarrow F_{update}(\alpha_{m})$ \\ 
			}
			$\beta_{r} = 0$ \\ \hfill {$\rightarrow$ Re-record average reward 
				values under new hyperparameters}\\
		}
		Perform soft-update of $\phi^{'}$ and $\theta^{'}$
	}
	\caption{Multi-Objective Soft Elastic Actor and Critic}
\end{algorithm}

The maximum number of training steps is denoted as $t_{max}$
\citep{sutton2018reinforcement}, while $k_{length}$ indicates the maximum number
of exploration steps per episode \citep{sutton2018reinforcement}. The initial
random exploration phase comprises $k_{init}$ steps
\citep{sutton2018reinforcement}. $k_{update}$ \citep{sutton2018reinforcement} is
the update interval determining the frequency of updates for these neural
networks used in the actor and critic policies. The reward $R_i$ is computed as
$R(S_i, A_i, D_i)$, where $D_i$ falls within the interval $[D_{min}, D_{max}]$,
representing the duration of the action.

Our algorithm employs a scalarizing strategy to optimize multiple objectives,
sweeping the design space by adapting $\alpha_m$ during training. Unlike
Hierarchical Reinforcement Learning (HRL)~\citep{dietterich2000hierarchical,
	li2019hierarchical} which aims for Pareto optimality~\citep{monfared2021pareto}
through layered reward policies, our technique simplifies the process. We focus
on ease of use and computational efficiency, ensuring our technique can easily
adapt to various algorithms.

It is important to note that the choice of $\psi$ is crucial for optimal performance,
as $\psi$ represents the sweeping step of our optimization (similar to the
learning parameter in gradient descent). We suggest utilizing the pre-set $\psi$
value from our implementation to minimize the need for further adjustments. If
training performance is inadequate, a high $\psi$ value might cause the reward
signal's gradient to change too quickly, leading to instability; in this case, a
lower $\psi$ is recommended. Conversely, if training progresses too slowly, a
low $\psi$ value might weaken the reward signal, hindering convergence; thus,
increasing $\psi$ could be beneficial.

\subsection{Convergence Proof}
\label{sec:convergence_analysis}

With our reward function, the policy gradient is:
\begin{equation}
\begin{aligned}
\nabla_\theta J(\pi_\theta) = \mathbb{E}_{\pi_\theta} \Big[ 
&\nabla_\theta \log \pi_\theta(a, D|s) \cdot \\
&\big( Q^\pi(s, a, D) \cdot (\alpha_{m} \cdot R_{\tau}) - 
\alpha_{\varepsilon} \big) \Big]
\end{aligned}
\end{equation}
where $\nabla_\theta J(\pi_\theta)$ is the gradient of the objective function 
with respect to the policy parameters $\theta$.

The value function update, incorporating the time dimension $D$ and our reward 
function, is:
\begin{equation}
\begin{aligned}
L(\phi) = \mathbb{E}_{(s, a, D, r, s')} \Big[ 
&\big( Q_\phi(s, a, D) - \big( r + \gamma \mathbb{E}_{(a', D') \sim \pi_\theta} \\
&[V_{\bar{\phi}}(s') - \alpha \log \pi_\theta(a', D'|s')] \big) \big)^2 \Big]
\end{aligned}
\end{equation}
where $L(\phi)$ is the loss function for the value function update, $r$ is the 
reward, $\gamma$ is the discount factor, and $V_{\bar{\phi}}(s')$ is the target 
value function.

The new policy parameter $\theta$ update rule is:
\begin{equation}
\begin{aligned}
\theta_{k+1} = \theta_k + \beta_k \mathbb{E}_{s \sim D, (a, D) \sim \pi_\theta} \Big[ 
&\nabla_\theta \log \pi_\theta(a, D|s) \cdot \\
&\big( Q_\phi(s, a, D) \cdot (\alpha_{m} \cdot R_{\tau}) - \alpha_{\varepsilon} \\
&- V_{\bar{\phi}}(s) + \alpha \log \pi_\theta(a, D|s) \big) \Big]
\end{aligned}
\end{equation}
where $\beta_k$ is the learning rate at step $k$.

To analyze the impact of dynamically adjusting $\alpha_{m}$ and $\alpha_{\varepsilon}$,
we assume the following conditions. The dynamic adjustment rules specify that
$\alpha_{m}$ increases monotonically by a small increment $\psi$ if the reward trend
decreases over consecutive episodes. Its upper limit, $\alpha_{max}$, guarantees
algorithmic convergence and prevents reward explosion. On the other hand,
$\alpha_{\varepsilon}$ decreases as defined.

The learning rate conditions must be satisfied, where $\alpha_k$ and $\beta_k$ adhere
to the following equations~\citep{konda1999actor}:
\begin{equation}
\sum_{k=0}^{\infty} \alpha_k = \infty, \quad \sum_{k=0}^{\infty} \alpha_k^2 < \infty
\end{equation}
\begin{equation}
\sum_{k=0}^{\infty} \beta_k = \infty, \quad \sum_{k=0}^{\infty} \beta_k^2 < \infty
\end{equation}

Assuming the critic estimates are unbiased:
\begin{equation}
\mathbb{E}[Q_\phi(s, a, D) \cdot (\alpha_{m} \cdot R_{\tau}) - \alpha_{\varepsilon}] = Q^\pi(s, a, D) \cdot (\alpha_{m} \cdot R_{\tau}) - \alpha_{\varepsilon}
\end{equation}
Since $R_{\tau}$ is a positive number within [0, 1], its effect on $Q^\pi(s, a, D)$ is
linear and does not affect the consistency of the policy gradient.

The positive scaling condition states that as $0 \leq R_{\tau} \leq 1$ and $\alpha_{m} 
\geq 0$, $\alpha_{m}$ only scales the reward without altering its sign. This scaling
does not change the direction of the policy gradient but affects its magnitude.
Additionally, a small offset $\alpha_{\varepsilon}$ is used to accelerate training.
This small offset does not affect the direction of the policy gradient but introduces a
minor shift in the value function, which does not alter the overall policy update
direction.

Under these conditions, MOSEAC will converge to a local optimum~\citep{sutton2018reinforcement}:
\begin{equation}
\lim_{k \to \infty} \nabla_\theta J(\pi_\theta) = 0
\end{equation}

\subsection{Lyapunov Stability}
\label{sec:lyapunov_function}

To analyze the stability of MOSEAC, we define a Lyapunov function $V(t)$ and calculate
its time derivative to ensure asymptotic stability:
\begin{equation}
V(t) = \frac{1}{2} \alpha_m(t)^2 + \sum_{s} \left[ Q(s, a, D, \alpha_m) 
- Q^*(s, a, D) \right]^2
\end{equation}
where $\alpha_m(t)$ varies with time, $Q(s, a, D, \alpha_m)$ is the current
Q-value ($D$ is the duration of action $a$), and $Q^*(s, a, D)$ is the ideal
Q-value.

The time derivative of $V(t)$ is:
\begin{equation}
\begin{split}
\dot{V}(t) = &\ \alpha_m(t) \dot{\alpha}_m(t) \\
& + \sum_{s} 2 \left[ Q(s, a, D, \alpha_m) - Q^*(s, a, D) \right] \\
& \dot{Q}(s, a, D, \alpha_m)
\end{split}
\end{equation}
Based on the linear growth model of $\alpha_m$:
\begin{equation}
\dot{\alpha}_m(t) = 
\begin{cases} 
k & \text{if } \alpha_m(t) < \alpha_{\text{max}} \\ 
0 & \text{if } \alpha_m(t) = \alpha_{\text{max}}
\end{cases}
\end{equation}
Replacing in $\dot{V}(t)$:
\begin{equation}
\dot{V}(t) = 
\begin{cases} 
\begin{aligned}
& \alpha_m(t) k + \sum_{s} 2 \left[ Q(s, a, D, \alpha_m) - Q^*(s, a, D) \right] \\
& \quad \dot{Q}(s, a, D, \alpha_m), \quad \text{if } \alpha_m(t) < \alpha_{\text{max}}
\end{aligned} \\ &  \\ 
\begin{aligned}
& \sum_{s} 2 \left[ Q(s, a, D, \alpha_m) - Q^*(s, a, D) \right] \\
& \quad \dot{Q}(s, a, D, \alpha_m), \quad \text{if } \alpha_m(t) = \alpha_{\text{max}}
\end{aligned} & 
\end{cases}
\end{equation}

Since $\alpha_m(t) > 0$ and $k > 0$, $\dot{V}(t) > 0$ when $\alpha_m(t) <
\alpha_{\text{max}}$. However, when $\alpha_m(t)$ reaches $\alpha_{\text{max}}$,
$\dot{V}(t) = 0$, indicating that the system reaches a stable state.

\section{Experimental Setup}
\label{sec:env}
\begin{figure}[!t]
	\centering
	\includegraphics[width=3.0in]{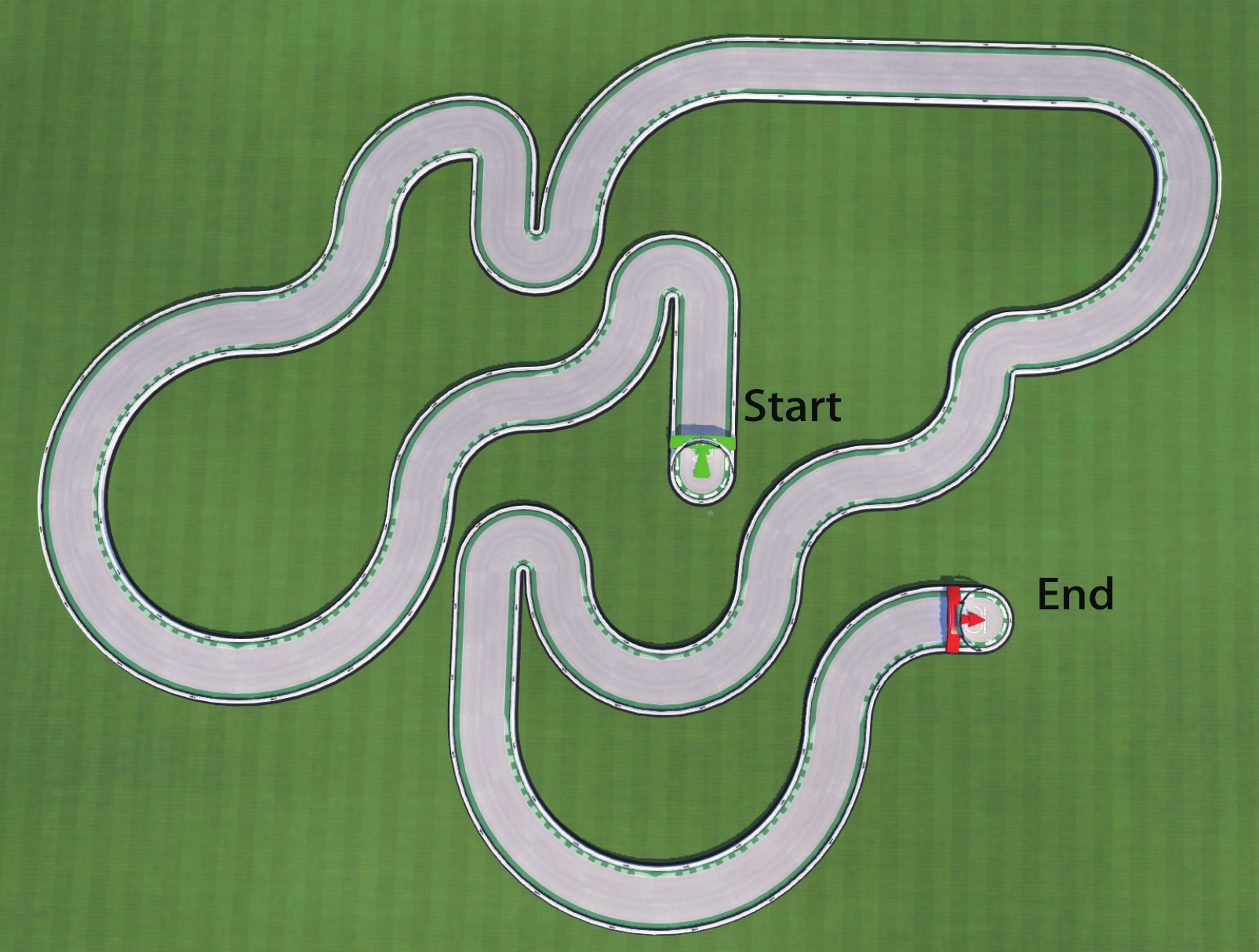}
	\caption{Top preview of the Trackmania track.}
	\label{fig_tmrlmap}
\end{figure}

We validate our MOSEAC in a real-time racing game, Trackmania~\citep{trackmania}.
Figure \ref{fig_tmrlmap} illustrates the testing environment. In Trackmania,
players race to complete the track as quickly as possible. We employed the map
developed by the TMRL~\citep{tmrl} team for a direct comparison of MOSEAC's
performance against their SAC model. It is important to note that training and
deploying a policy for this game can only happen in real
time~\citep{ramstedt2019real}, ensuring the realism of the overall experiments.

For controlling the car in the game, Trackmania offers:
\begin{inparaitem}
	\item gas control;
	\item brake control; and
	\item steering.
\end{inparaitem}
The action value includes the rate at which these controls are applied. The
input to the policy is the pixels as well as the numerical information shown on
the heads-up display.

The reward system in the game is designed to encourage efficient path following:
more rewards are given for covering more path points in a single move.This path 
is a series of evenly distributed points that collectively form the shortest 
route from the starting point to the end point of the racing game. This 
technique is consistent with the TMRL team's methodology~\citep{tmrl}.
The game uses realistic car physics, but there is no car damage or crash
detection. Further details are available in Table \ref{table:tmrl_info}.

\begin{table}[htbp]
	\centering
	\caption{Trackmania Environments Details}
	\resizebox{\columnwidth}{!}{
		\begin{tabular}{lll}
			\toprule
			\multicolumn{3}{c}{State and action space of Trackmania} \\
			\midrule
			& Data Space                        & Annotate   \\
			\midrule
			State Dimension       & $143$           & Details in Section \ref{fig_Seac_str_trackmania}  \\
			Car Speed                  & $(-1, 1)$           &        \\
			Car Gear    & $(-1, 1)$           &       \\
			Wheel RPM & $(-1, 1)$                   &               \\
			RGB Image & $(64, 64, 3)$                   & RGB arrays              \\
			Action Dimension       & $4$           &  \\
			Gas Control         & $(-1, 1)$          &                     \\
			Brake                & $(-1, 1)$          &                    \\
			Yaw Control       & $(-1, 1)$          &                    \\
			Control Rate          & $(5, 30)$ $Hz$        &                   \\
			\bottomrule
	\end{tabular}}
	\label{table:tmrl_info}
\end{table}

%

\begin{figure*}[!t]
	\centering
	\includegraphics[width=5.2in]{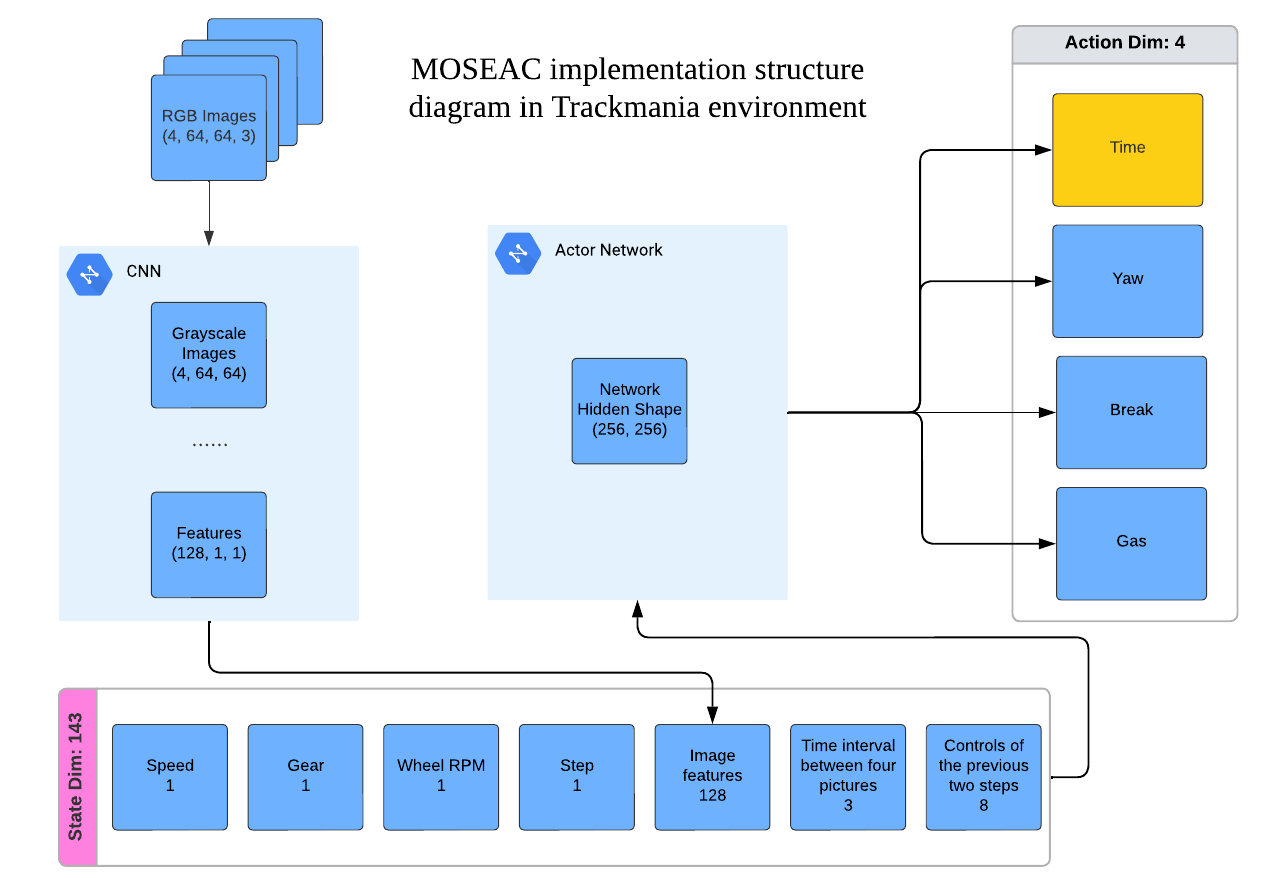}
	\caption{This is the implementation structure diagram of MOSEAC in the
		TrackMania environment. We use CNN to extract potential information in the
		environment and learn the extracted feature values based on rewards. The
		143-dimensional state value and the 4-dimensional action value are shown in
		the figure. The time in the action value is not used for the current time
		step but for the next step.}
	\label{fig_Seac_str_trackmania}
\end{figure*}

To facilitate the agent's comprehension of how the controls influence speed and
acceleration, we feed 4 sequential frames (and the intervals between them) to a
convolutional neural network (CNN), whose output is an embedding contributing to
the state representation (see Figure~\ref{fig_Seac_str_trackmania}). The CNN
distills features, transforming the image data from a matrix of dimensions (4,
64, 64, 3) to a compact (128, 1, 1) matrix.

To evaluate the impact of our variable time step technique, we used the same
visual-based navigation and task reward policy across the tested approaches,
varying only the control rate. It is important to note that the task reward policy
only includes time and disregards collisions.


Overall, our model inputs are the car's speed, gear, and wheel RPM, the current 
step number in one episode, along with the two most recent actions taken, for a 
final 143-dimensional state.

\begin{figure}[!t]
	\centering
	\includegraphics[width=\columnwidth]{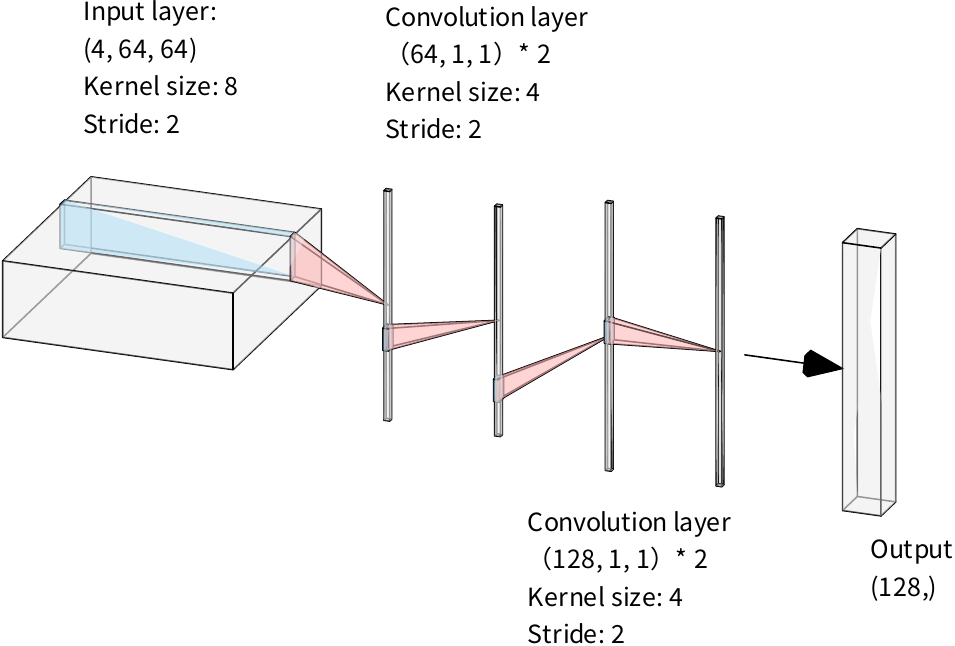}
	\caption{Our CNN structure diagram that we used to extract 
		image features from the Trackmania video game. We convert RGB images 
		into grayscale images and then input them into the CNN.}
	\label{fig_cnn}
\end{figure}

\section{Experimental Results}
\label{sec:results}

We conducted experiments with MOSEAC, CTCO~\citep{karimi2023dynamic} and
SEAC~\citep{wang2024deployable} on Trackmania for over 1320 hours\footnote{Our
	code is publicly available on GitHub:
	\url{https://github.com/alpaficia/TMRL_MOSEAC}}.
These experiments were conducted on a I5-13600K computer with an NVIDIA RTX 4070
GPU. The final result video for MOSEAC is publicly
available\footnote{\url{https://youtu.be/1aQ0xSK55nk}}, with 43.202 seconds to
complete the test track.


\begin{figure}[!t]
	\centering
	\includegraphics[width=\columnwidth]{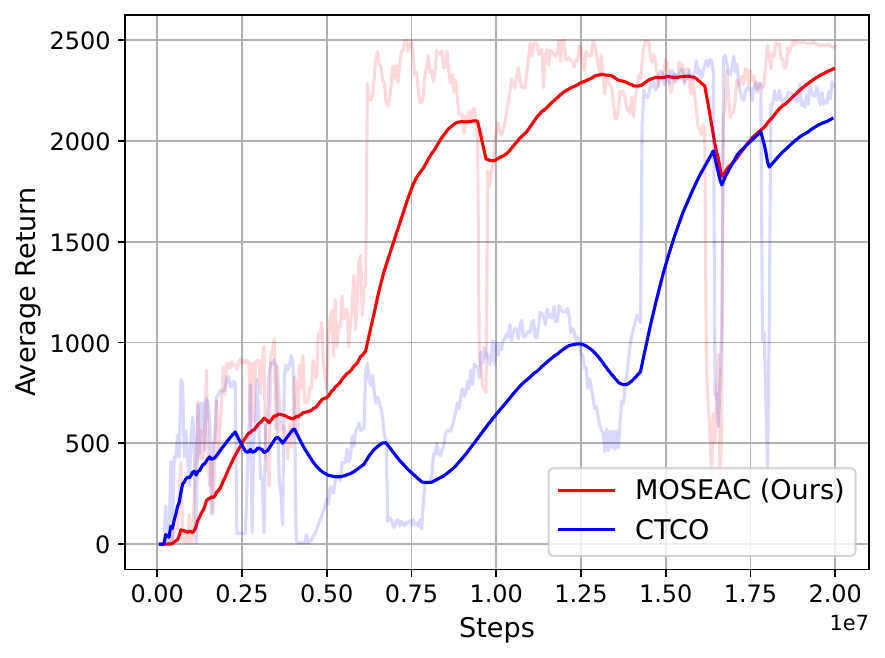}
	\caption{Training Progress of MOSEAC and CTCO: Average Return Over Time}
	\label{fig_return_ral}
\end{figure}

\begin{figure}[!t]
	\centering
	\includegraphics[width=\columnwidth]{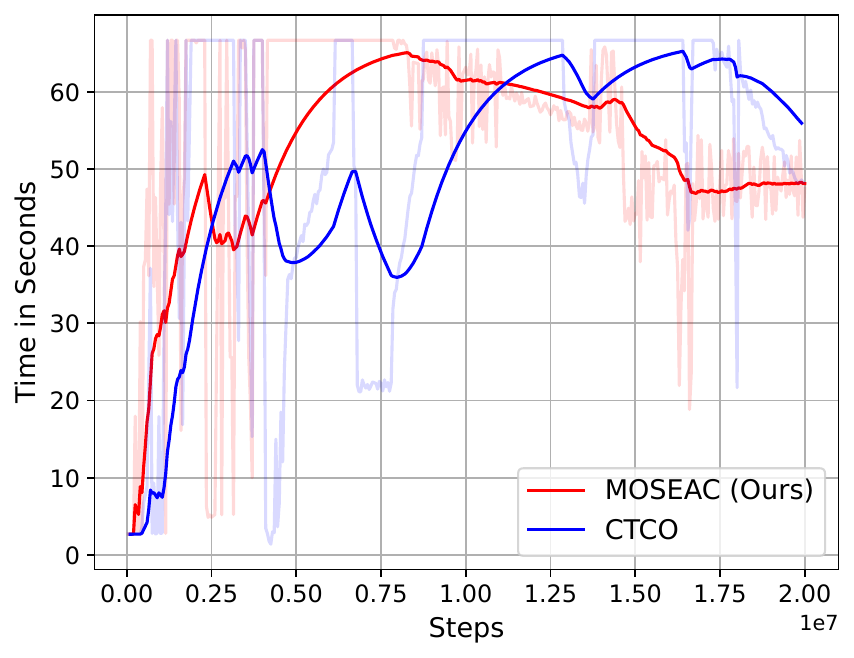}
	\caption{Training Progress of MOSEAC and CTCO: Time Consumption Over Steps}
	\label{fig_tr_speed_ral}
\end{figure}

Before discussing the performance of MOSEAC and CTCO, it is essential to provide
some background context. In our previous work, the SEAC model policies were
trained utilizing different hyperparameters~\citep{wang2024reinforcement}. These
hyperparameters were explicitly tuned for SEAC, optimizing its performance in
the given environment. Besides, the SAC model was trained by the TMRL team with
a different set of hyperparameters tailored to their training
technique~\citep{tmrl}. Given these differences, directly comparing their training
curves would be unfair; hence, we focus solely on the MOSEAC and CTCO training
curves.

Reward signals are notably sparse in the environment set by the TMRL team. The
reward is calculated based on the difference between the path coordinates after
movement and the initial path coordinates, divided by 100~\citep{tmrl}. This
sparse reward environment necessitates a very low-temperature coefficient alpha
SAC~\citep{haarnoja2018soft1}, otherwise, the entropy component
would dominate the reward. This insensitivity to reward signals can lead to poor
training performance or even training failure. MOSEAC and CTCO are derived from
SAC~\citep{wang2024moseac, karimi2023dynamic}, and they face similar constraints.

We have maintained this reward system to ensure a fair comparison with the TMRL
team. Modifying the environment is not straightforward, as simply amplifying the
reward signal may cause the agent to accumulate rewards rapidly, leading to
overly optimistic estimates and unstable training. Such ``reward explosion'' can
cause the agent to settle into local optima, neglecting long-term returns and
reducing overall performance.

Using an adaptive temperature coefficient during training, we observed that the
coefficient value becomes extremely low towards the end of the training. This
results in a narrow distribution of actions, significantly slowing down the
policy training and optimization process. This small temperature coefficient
causes both MOSEAC and CTCO to have a slow training and optimization process.

\autoref{fig_return_ral} demonstrates that MOSEAC reaches a high point at around
6 million steps, indicating initial training success. However,
\autoref{fig_tr_speed_ral} shows that after 6 million steps, MOSEAC required
over 14 million additional steps to technique the optimal value. This indicates
that MOSEAC and CTCO require substantial time for initial training and further
optimization in a sparse reward environment.

Despite these challenges, MOSEAC exhibits better training efficiency than CTCO.
MOSEAC adapts to the environment more quickly, maintaining a more stable
learning curve. In contrast, CTCO, influenced by its adaptive $\gamma$
mechanism, tends to favor smaller $\gamma$ values, impairing long-term planning
and significantly slowing down training. Furthermore, MOSEAC achieves a higher
final reward than CTCO, showcasing its superior adaptability and efficiency in
sparse reward settings.


Training in the Trackmania environment is real-time, and our 1320+
hours of training have shown that MOSEAC can handle complex, sparse
reward environments, albeit requiring notable time investment. Therefore, if
a direct comparison with our results, specifically regarding training speed
within the same reward environment, is unnecessary, we recommend optimizing the
reward signals in the TMRL Trackmania environment. One possible optimization
strategy could be to amplify the per-step reward signal appropriately. This
adjustment theoretically allows for faster training of effective control
policies, significantly enhancing training efficiency and saving substantial
real-world time.


\begin{figure}[!t]
	\centering
	\includegraphics[width=\columnwidth]{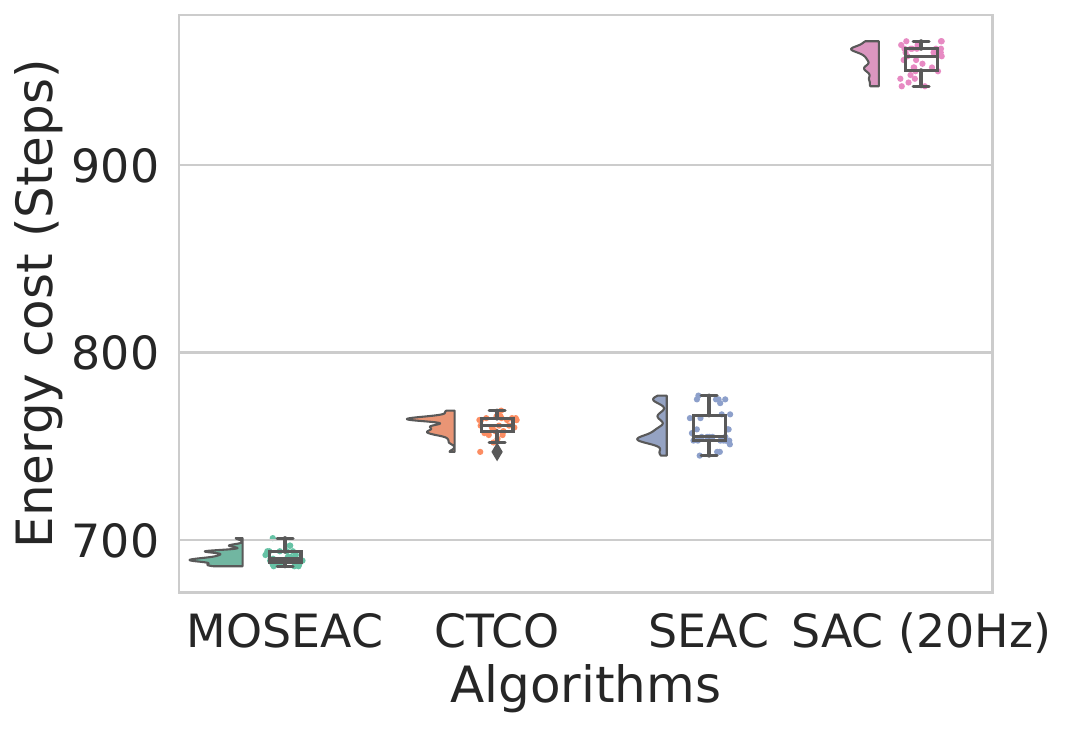}
	\caption{This figure compares the energy costs of four different 
		algorithms: MOSEAC, CTCO, SEAC, and SAC (20Hz). The y-axis represents 
		the energy cost in steps, while the x-axis lists the algorithms. The 
		box plots demonstrate that MOSEAC has the lowest median energy cost, followed 
		by SEAC, CTCO, and SAC (20Hz). This indicates that MOSEAC is the most 
		energy-efficient among the four algorithms.}
	\label{fig_energy}
\end{figure}

\begin{figure}[!t]
	\centering
	\includegraphics[width=\columnwidth]{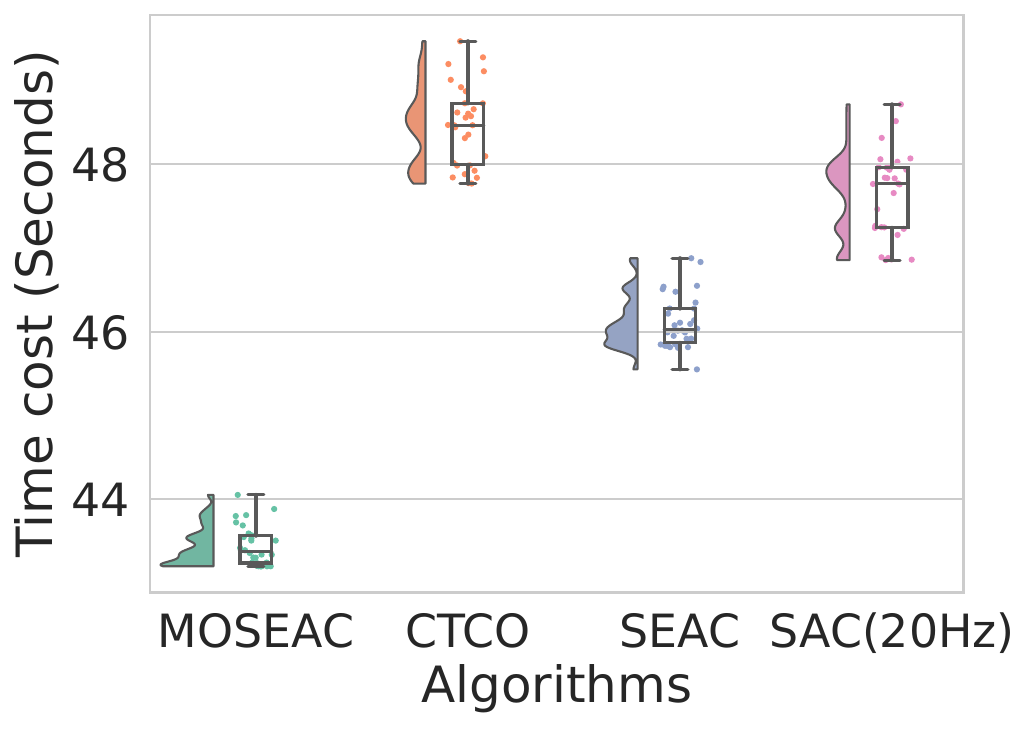}
	\caption{This figure compares the time costs of four different algorithms: 
		MOSEAC, CTCO, SEAC, and SAC (20Hz). The y-axis represents the time cost 
		in seconds, while the x-axis lists the algorithms. The box plots reveal 
		that MOSEAC has the lowest median time cost, indicating faster task 
		completion compared to the other algorithms. CTCO, SEAC, and SAC (20Hz) 
		demonstrate higher median time costs, with MOSEAC outperforming them in terms 
		of time efficiency.}
	\label{fig_time_cost}
\end{figure}

We compared the racetrack time (referred to as \emph{time cost}) as well as the
computational energy cost (in terms of the number of control steps) of MOSEAC
with SEAC, CTCO, and SAC (20Hz) after training. \autoref{fig_energy} illustrates
the energy cost distribution and \autoref{fig_time_cost} the time cost (lower is
better in both cases). For both energy and time there is no overlap among the
different methods, and the data follows a normal distribution (Shapiro-Wilk test
of normality, see \autoref{tab:testOfNormality(Shapiro-Wilk)_energy_ral} and
\autoref{tab:testOfNormality(Shapiro-Wilk)_time_ral}).

Using a paired sample T-test, MOSEAC showed lower energy in all of the trials
compared with CTCO ($t=-64.85$, $df=29$, $p \ll 0.001$), SEAC ($t=-35.90$,
$df=29$, $p\ll 0.001$), and SAC ($t=-196.35$, $df=29$, $p \ll 0.001$). The statistics for the
energy cost measures are presented in \autoref{tab:descriptives_energy_ral}.
Similarly, MOSEAC showed lower race time in all of the trials compared with CTCO
($t=-55,67$, $df=29$, $p \ll 0.001$), SEAC ($t=-32.92$, $df=29$, $p \ll 0.001$), and SAC
($t=-41.06$, $df=29$, $p \ll 0.001$). The statistics for the
time cost measures are presented in \autoref{tab:descriptives_time_ral}.



\begin{table}[h]
	\centering
	\caption{Test of Normality (Shapiro-Wilk) - Energy cost difference}
	\label{tab:testOfNormality(Shapiro-Wilk)_energy_ral}
	\begin{tabular}{lrrrr}
		\toprule
		& & & W & p \\
		\cmidrule[0.4pt]{1-5}
		MOSEAC & - & CTCO & 0.947 & 0.142 \\
		MOSEAC & - & SEAC & 0.948 & 0.150 \\
		MOSEAC & - & SAC (20Hz) & 0.956 & 0.242 \\
		\bottomrule
	\end{tabular}
\end{table}


\begin{table}[h]
	\centering
	\caption{Energy cost descriptives}
	\label{tab:descriptives_energy_ral}
	\resizebox{\columnwidth}{!}{
		\begin{tabular}{lrrrrr}
			\toprule
			& N & Mean & SD & SE & COV \\
			\cmidrule[0.4pt]{1-6}
			MOSEAC & 30 & 690.800 & 3.652 & 0.667 & 0.005 \\
			CTCO & 30 & 760.933 & 4.989 & 0.911 & 0.007 \\
			SEAC & 30 & 759.467 & 9.508 & 1.736 & 0.013 \\
			SAC (20Hz) & 30 & 956.133 & 7.482 & 1.366 & 0.008 \\
			\bottomrule
	\end{tabular}}
\end{table}

\begin{table}[h]
	\centering
	\caption{Test of Normality (Shapiro-Wilk) - Time cost difference}
	\label{tab:testOfNormality(Shapiro-Wilk)_time_ral}
	\begin{tabular}{lrrrr}
		\toprule
		& & & W & p \\
		\cmidrule[0.4pt]{1-5}
		MOSEAC & - & CTCO & 0.954 & 0.222 \\
		MOSEAC & - & SEAC & 0.963 & 0.367 \\
		MOSEAC & - & SAC (20Hz) & 0.968 & 0.478 \\
		\bottomrule
	\end{tabular}
\end{table}


\begin{table}[h]
	\centering
	\caption{Time cost descriptives}
	\label{tab:descriptives_time_ral}
	\resizebox{\columnwidth}{!}{
		\begin{tabular}{lrrrrr}
			\toprule
			& N & Mean & SD & SE & COV \\
			\cmidrule[0.4pt]{1-6}
			MOSEAC & 30 & 43.441 & 0.237 & 0.043 & 0.005 \\
			CTCO & 30 & 48.463 & 0.481 & 0.088 & 0.010 \\
			SEAC & 30 & 46.117 & 0.317 & 0.058 & 0.007 \\
			SAC (20Hz) & 30 & 47.636 & 0.510 & 0.093 & 0.011 \\
			\bottomrule
	\end{tabular}}
\end{table}

The improved performance of MOSEAC over SEAC, CTCO, and SAC (20Hz) can be 
attributed to its reward function and the integration of a state variable. 
SEAC uses a linear reward function that independently combines task 
reward, energy penalty, and time penalty. In contrast, MOSEAC employs a 
multiplicative relationship between task and time-related rewards. This 
non-linear interaction enhances the reward signal, especially when task 
performance and time efficiency are high, thereby naturally balancing 
these factors. By keeping the energy penalty separate, MOSEAC retains 
flexibility in tuning without complicating the relationship between time 
and task rewards. This design allows MOSEAC to guide the agent's decisions 
more effectively, resulting in improved energy efficiency and faster task 
completion in practical applications.


\section{Conclusions}
\label{sec:conclusions}
In this paper, we introduced and evaluated the Multi-Objective Soft Elastic
Actor-Critic (MOSEAC) algorithm, demonstrating its superior performance compared
to CTCO, SEAC, and SAC (20Hz) in the sparse reward environment created by the
TMRL team in the Trackmania game. MOSEAC features an innovative reward function
that combines task rewards and time-related rewards, enhancing the reward signal
when both task performance and time efficiency are high. This non-linear
interaction balances these factors, leading to improved energy efficiency and
task completion speed.

To prevent reward explosion, we introduced an upper limit ($\alpha_{max}$) for
the hyperparameter $\alpha_m$, ensuring convergence and stability, validated
utilizing a Lyapunov model. The boundedness analysis confirmed that $\alpha_m$
remains within the defined limits, maintaining the strength of the reward
function and ensuring the robustness and reliability of the learning process.

Our experiments in the Trackmania environment, conducted over 1300 hours,
validated these enhancements, showcasing MOSEAC's robustness and efficiency in
real-world applications. The recorded best time for MOSEAC is 43.202 seconds,
significantly faster than the other algorithms.

Building on MOSEAC's success, our future work will focus on extending the
principles of variable time step algorithms to other RL
frameworks, such as Hierarchical Reinforcement Learning (HRL). This extension
aims to address the complexities of long-term planning tasks more efficiently,
further improving the adaptability and performance of RL algorithms in diverse
and challenging environments, which brings benefits for solving complex problems
by RL utilizing real robot systems.

\bibliographystyle{elsarticle-harv}
\bibliography{reference.bib}

\subsection*{}
\setlength\intextsep{0pt} 
\begin{wrapfigure}{l}{25mm}
	\centering
	\includegraphics[width=1in,height=1.25in,clip,keepaspectratio]{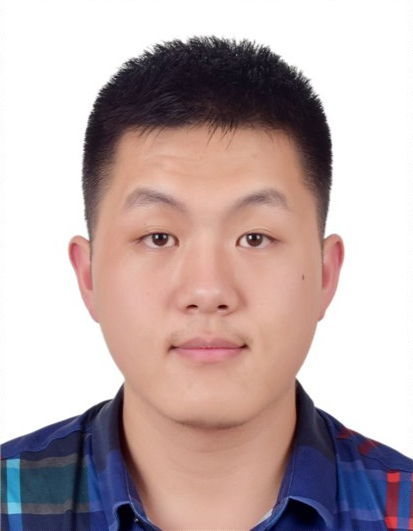}
\end{wrapfigure}
\noindent \textbf{Dong Wang} received his bachelor's degree in electronic engineering from the 
School of Aviation, Northwestern Polytechnical University (NWPU), Xi'an, China, 
in 2017. He is pursuing his Ph.D. in the Department of Software Engineering at 
Polytechnique Montreal, Montreal, Canada. His research interests include 
RL, computer vision, and robotics.

\hspace*{\fill} 

\subsection*{}
\setlength\intextsep{0pt} 
\begin{wrapfigure}{l}{25mm}
	\centering
	\includegraphics[width=1in,height=1.25in,clip,keepaspectratio]{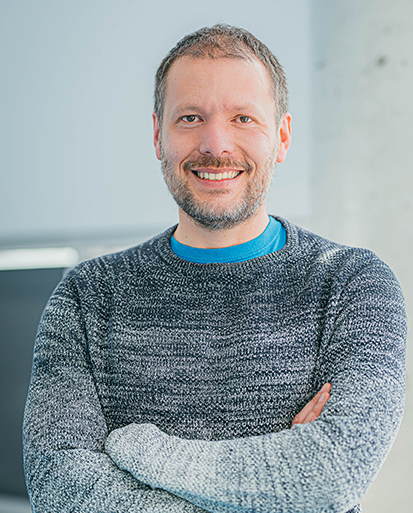}
\end{wrapfigure}
\noindent \textbf{Giovanni Beltrame} eceived
the Ph.D. degree in computer engineering from Po-
litecnico di Milano, Milan, Italy, in 2006.
He worked as a Microelectronics Engineer with the

\end{document}